**A Multi-Graph Convolutional Neural Network Model for Short-Term Prediction of Turning Movements at Signalized Intersections**

**Jewel Rana Palit***
Graduate Research Assistant
Department of Civil and Chemical Engineering
University of Tennessee
430A EMCS Building
615 McCallie Ave, Chattanooga, TN 37403
Email: jnc411@mocs.utc.edu

**Osama A. Osman*, Ph.D. (Corresponding author)**
ORCiD: https://orcid.org/0000-0002-5157-2805
Assistant Professor
Department of Civil and Chemical Engineering
University of Tennessee
430A EMCS Building
615 McCallie Ave, Chattanooga, TN 37403
Tel: 423-425-4398 | Fax: 423-425-5229
Email: osama-osman@utc.edu

*Equally Contributed to the work

**ABSTRACT**

Traffic flow forecasting is a crucial first step in intelligent and proactive traffic management. Traffic flow parameters are volatile and uncertain which makes traffic flow forecasting a hard task to perform if the right forecasting model is not used. Additionally, the non-Euclidian data structure of traffic flow parameters is challenging to analyze both from a spatial and temporal perspective. State of art deep learning approaches adopts pure convolution, recurrent neural network, and hybrid method to achieve this objective efficiently. However, many of the approaches in the state-of-the-art literature rely on complex architectures that could make them hard to train. This in addition to the fact that such complex architecture could add to the black box nature of deep learning. This study introduces a novel deep learning architecture, referred to as multigraph convolution neural network (MGCNN), for turning movement prediction at intersections. The proposed architecture combines a multigraph structure, built to model temporal variations in traffic data, with a spectral convolution operation to support modeling the spatial variations in traffic data over the graphs. The proposed model was tested using a twenty-day flow and traffic control data collected from an arterial in downtown Chattanooga, TN, with 10 signalized intersections. The model's ability to perform short-term predictions over 1, 2, 3, 4, and 5 minutes into the future was evaluated against four baseline state-of-the-art models. The results showed that our proposed model is superior to the other baseline models in predicting turning movements with an MSE of 0.9.

**INTRODUCTION**

The computational advancement in Intelligent Transport System (ITS) has been among the core foundations for smart cities. Enabled traffic flow forecasting, analysis, and modeling through such advancements have always been integral parts of management of the multifaceted transportation systems. Traffic flow is however a non-linear and volatile concept that is subjected to dynamic changes in multiple parameters such as signal control, roadway geometric characteristics, interactions of other road users, and weather conditions, among others. Constant changes in these parameters make traffic flow more unpredictable. Moreover, the traffic flow dynamically changes over time with or without any specific pattern which is why it has been more challenging to forecast traffic flow parameters in short-term or long-term horizons. While different parameters like travel time, speed, and mid-block traffic flow contribute equally to corridor-level traffic streams, directional traffic movements could be more challenging as they affect network-level traffic conditions. Efficient forecasting of such directional movements over is key to enable adaptive control at intersections, proactive traffic management decisions, and routing planning. Accurate prediction of these movements could be troublesome due to their variability both spatially and temporally. This study presents an approach to predict turning movements at intersections through the application of deep learning.

  Because of the dynamic fluctuation in the traffic stream over time, it has always been the key attention of scholars to extract latent traffic patterns through the application of traditional and innovative machine learning approaches. Classical statistics and data-driven prediction methods are the two prediction approaches observed in the literature [1, 2]. Specifically, the autoregressive integrated moving average method which is also known as ARIMA [3], and its many modifications such as subset ARIMA [4], seasonal ARIMA [5], combined Kohonen maps-KARIMA [6], and Kalman Filtering method [7] are among the earliest noticed approaches for traffic flow prediction. However, these classical methods assume linearity in time sequences and thus cannot capture the non-linear nature of traffic flow. Therefore, researchers used more innovative machine learning approaches such as Support Vector Regression (SVR) [8] based methods and k-nearest neighbor models [9] as an attempt to overcome e the limitations of classical statistics methods. Nonetheless, many machine learning approaches are yet considered very shallow, making them unable to capture the volatile nature and uncertainty of traffic flow, which makes them in need of custom-made data features [10]. Additionally, these models may not be able to capture the spatial-temporal aspects of traffic flow at a time [11].

  Recently, deep learning methods have been continuously emerging and developing in the field of Intelligent Transport Systems. Scholars applied deep learning approaches to address the limitations of classical statistics and traditional machine learning methods and have been able to improve their model's performance drastically in traffic prediction while dealing with big data. For instance, a deep belief network (DBN) was proposed by Huang et al. for traffic flow prediction which applied to multitask learning of the model [12]. Additionally, a Stacked Autoencoder (SAE) model was proposed using a layer-wise greedy algorithm to exploit spatial-temporal correlations in data to help improve flow prediction accuracy [13]. However, these approaches do not accurately extract the complex spatial-temporal correlations in traffic flow and have low prediction performance [10]. In essence, innovative deep learning approaches such as Convolutional Neural Network (CNN) and long short-term memory (LSTM) have been applied both separately and in a hybrid form to capture the spatial-temporal dynamics of traffic flow. Ma et al. proposed a CNN-based model to predict traffic speed. Spatial-temporal traffic dynamics were converted into images and a 2-D time-space matrix was used to express the correlations in time and space [14]. However, this model was not efficient enough as traffic data is non-Euclidean while CNN works well with regular grided and Euclidean data structures [11]. To predict the time-series data nature of traffic flow, Zhao et al proposed LSTM based model using OD (origin-destination) matrix to correlate different links over the road network [15]. A link-level LSTM model was proposed to explore the irregular nature of travel time over 66 links in the England Highway System [16]. Similarly, Hou et al attempted to apply a stacked autoencoder with a Radial basis function (RBF) to predict short-term traffic flow while considering weather factors [17]. Their model also has an LSTM component to capture temporal correlations in traffic data. While these models do well in capturing temporal correlations, they do not completely understand the spatial correlations in the traffic flow over an entire network [11]. Mahmoud et al. predicted turning movements (left turn and thorough) for North and South Bound at cycle level using LSTM, GRU & Boosting method separately with different accuracies [18]. However, they still depended on RNN elements which are computationally heavy and not enough for capturing spatial-temporal correlation altogether. In an attempt to capture spatial-temporal correlations, Wu et al proposed a hybrid model which fused a 1-D CNN to capture spatial features and two LSTMs to capture the temporal characteristics of traffic flow [19].

While CNN and LSTM proved successful for traffic state prediction, they may not accurately capture data dependencies especially over space since a model like CNN is not designed for the non-grid data structure. Graph Neural Network (GNN) on the other hand has the ability to provide an accurate representation of the spatial information over a transportation network. In fact, transportation networks/corridor can be modeled as a directed or in-directed graph through GNNs. Thus, research focus has been directed to investigating how GNNs could help capture both temporal and spatial correlations in traffic data as well as model uncertainties in transportation systems for more accurate traffic prediction. As an example, a Diffusion Convolutional Recurrent Neural Network (DCRNN) was proposed by Li et al. for application over a directional graph representation of a transportation network for traffic speed prediction [20]. In another study, Zhao et al proposed a Temporal Graph Convolution Network (T-GCN) which used GCN with the Gated Recurrent Unit (GRU) to capture the spatial and temporal dependences separately [21]. To better capture the complex relationship of traffic data structure, Wang et al. proposed a hybrid model Spatial-Temporal Graph Neural Network(STGNN) framework which introduced a transformer layer with GCC and GRU [22]. Zhu et al. first used brief rule base (BRB)to fuse data and predict traffic speed [2]. Lv et al. introduced the Multigraph concept known as Temporal Multi-Graph convolutional neural network (TMGCNN) considering spatial, temporal, and semantic correlation [11]. TMGCNN works with multigraph constructed from three aspects mainly: road graph for topological structure, traffic patter correlation and local area functionality and convolute them to extract meaningful spatial information. Lu et al. combined LSTM and GNN in their proposed Graph Long Short-Term Memory Model (GLSTM) where they feed the LSTM an unweighted graph data. The message passing operation of GNN in their model is applied on it to capture spatial-temporal component and predict traffic speed [23]. Ni et al. used spatial-temporal blocks to stack multigraphs and proposed spatiotemporal gated multi-graph network (STGMN)[24]. Table 1 summarizes the comparative factors of traffic flow prediction approaches over the years.

*Palit and Osman***Table 1: Summary of Traffic Flow Prediction Approaches from the Literature**

| Authors | Model | Model Nature | Predicted Parameter | Helpful Aspect | Drawbacks |
|---|---|---|---|---|---|
| **Hamed et al.,1995** | ARIMA | Classic Statistical | Traffic volume | Easy implementation. | Cannot predict the non-linear nature of traffic flow. |
| **Wu et al.,2004** | SVR | Data-driven machine learning model. | Travel time | Outperform traditional classical statistical methods. | Cannot capture spatial correlation. |
| **Huang et al., 2014** | DBN | Deep Learning | Traffic volume | First attempt using deep learning for traffic flow prediction. | Fail to extract complex spatiotemporal features. |
| **Ma et al.,2017** | Deep Convolutional Neural Network model. | CNN | Traffic speed | Perform well while capturing spatial information. | Only Suitable for Euclidian data while traffic data points are non-Euclidian. |
| **Mahmoud et al., 2021** | Recurrent Neural Network | LSTM, GRU | Turning Movement | Capture temporal components of non-grided traffic data structure. | Less efficient in capturing spatial correlation. |
| **Li et al.,2017** | DCRNN | Hybrid | Traffic speed | Obtain both spatial and temporal efficiency using GCN and RNN. | Heavy RNN computation. |
| **Zhao et al.,2019** | T-GCN | Hybrid | Traffic speed | Obtain both spatial and temporal efficiency using GCN and RNN. | Computationally exhausting and heavy RNN computation. |
| **Zhu et al.,2020** | Belief Rule Based (BRB) RNN-GCN model | Hybrid | Traffic speed | Multiple weather factors affecting traffic stream were considered. | Heavy computation of RNN and expensive. |
| **Lv et al,2020** | TMGCNN | Hybrid | Traffic speed | Multi graph layer extract spatial information from broader horizon. | Complex architecture and exhausting. |
| **Ni et al., 2022** | STGMN | GNN | Traffic flow | Does not require RNN element | Uses spatial temporal blocks which are slow while processing large amount of data and susceptible to overfitting. |

**Research Gaps and Contributions**
The literature summary shows that significant effort was done to achieve efficient traffic forecasting from both spatial and temporal points of view. While the application of GNN proved to be superior compared to other models, to our knowledge most of the GNN-based models were associated with complex architecture that incorporated hybrid models to extract the spatial-temporal correlations in the data. Few multigraph concepts have been observed in studies like DCRNN and T-MGCNN, but they still had RNN incorporated to learn in temporal dimension. Such hybrid architectures are usually associated with difficulties in training and require heavy computations [1], this is in addition to adding to the black-box nature of deep learning models. Recurrent networks are well known to need iterative training, which involves error accumulation by steps, for sequence learning. Also, the gradient explosion or gradient vanish issue, affects the RNN, hinders full time series learning and information is lost as time sequence grows [25, 26]. Advanced multigraphs method like STGMN used spatial temporal blocks to get rid of RNN element. However, they are computationally intensive while processing large amount of data and easily susceptible to overfitting [27, 28]. Also, majority of these models follow spatial approaches for convolution operation. These struggle to scale to large graphs and has limited ability to handle irregular non-Euclidean graph structures [29] and traffic data suits this type perfectly.

In essence, this study proposes to capitalize on the benefits of GNNs by developing a multigraph convolution architecture to model temporal and spatial dependencies in traffic data and predict turning movements at intersections. Our proposed architecture, multigraph convolution neural network (MGCNN), employs a pure convolutional structure combining GNN for capturing spatial information and convolution on time to obtain temporal dynamics. While previous GNN and multigraph approaches relied on RNN element to model temporal components, we get rid of this need of RNN element saving us from heavy computation, model complexity and training time ignoring information loss due to gradient issues at the same time. MGCNN also does not require temporal blocks which makes it suitable for irregular traffic data and less prone to overfitting. It uses pure spectral convolution, which is robust, computationally effective, captures global graph structure and perfect for long-range dependencies [30-33]. This enables us to model spatial-temporal component of traffic dynamics effectively using multi-graph structures. We compare our model with other established approaches in the literature for further evaluation of performance efficiency.

**METHODOLOGY**

Our proposed Multigraph Convolutional Neural Network (MGCNN) consists of four main components: 1) multigraph construction (input) layer, which to define the number of graphs in the model input based on the time step size and the size of the lookback window – information on the nodes ($\mathcal{Y}_t^i \forall\ i = 1{:}n, n = number\ of\ nodes$) and edge weights $W_t^{i,j} \forall\ i = 1{:}n-1\ \&\ j = 2{:}n$) on each graph are also defined; 2) the multigraph stacking and fusion layer, which aims to fuse the constructed graphs over the different time steps to model the temporal pattern of the network state; 3) Graph Convolution Layers which convolute through the constructed multigraph with a sliding lookback window to extract the spatial correlation a predefined time window; 4) a fully connected layer to extract the desired output. Figure 1 represents the proposed MGCNN architecture.

**Multi-Graph Construction**
To predict turning movements at intersections, we define a transportation corridor by a set of bidirectional graphs where every graph represents the transportation network state at a given time. At any time step t, the graph is denoted by $\mathcal{G}_t = (\mathcal{V}, \mathcal{E}, W_t)$ where $\mathcal{V} = \{\mathcal{V}_1, \mathcal{V}_2, \ldots \mathcal{V}_n\}$, is a set of a finite number of vertices or nodes representing the intersections in the geographical area along a traffic corridor. $\mathcal{E}$ is the set of edges connecting the nodes which is the set of links between intersections. The weights of these connections are defined for every time step and are expressed by the weighted adjacency matrix $W_t \in \mathbb{R}^{n \times n}$ while $W_t$ is composed of the weights $w_t^{i,j}$, at time $t$, of the connection between every node $\mathcal{V}_i$ and node $\mathcal{V}_j$. If two nodes aren't connected the weight $w_t^{i,j} = 0$. $W_t$ can be represented as:

$$W_t = \begin{bmatrix} 0 & w_t^{1,2} & \cdots & w_t^{i,j} \\ w_t^{2,1} & 0 & \cdots & \vdots \\ \vdots & \vdots & \vdots & \vdots \\ w_t^{j,i} & \cdots & \cdots & 0 \end{bmatrix}$$

At any time t, the information at a node $i$ is defined as $\mathcal{Y}_t^i$, where $\mathcal{Y} = (\mathcal{Y}_1, \mathcal{Y}_2, \mathcal{Y}_3, \ldots \mathcal{Y}_\mathcal{K})$ is a set of features defining the state of that node at time *t*. Based on the defined $\mathcal{Y}_t^i$ array and $W_t^{i,j}$ matrix, a set of graphs $\{\mathcal{G}_{t-M}, \ldots \mathcal{G}_{t-3}, \mathcal{G}_{t-2}, \mathcal{G}_{t-1}, \mathcal{G}_t\}$ is constructed to represent the transportation network/corridor state over time window *t:t-M*, where *M* represents the lookback window over which the temporal correlations in data should be captured. For every graph defined at time *t*, a $\mathcal{Y}_t^i$ array and $W_t^{i,j}$ matrix is also defined.

**Fusion & Graph Stacking**
This model component aims to stack the graph sequence defined in the multigraph construction step to construct an ensemble of graphs that can help not only capture spatial correlations in traffic data but also models the temporal variations in the traffic state over the transportation network. This Multigraph structure is expressed as

$$G = \{\mathcal{G}_{t-M}, \ldots \mathcal{G}_{t-3}, \mathcal{G}_{t-2}, \mathcal{G}_{t-1}, \mathcal{G}_t\} \tag{1}$$

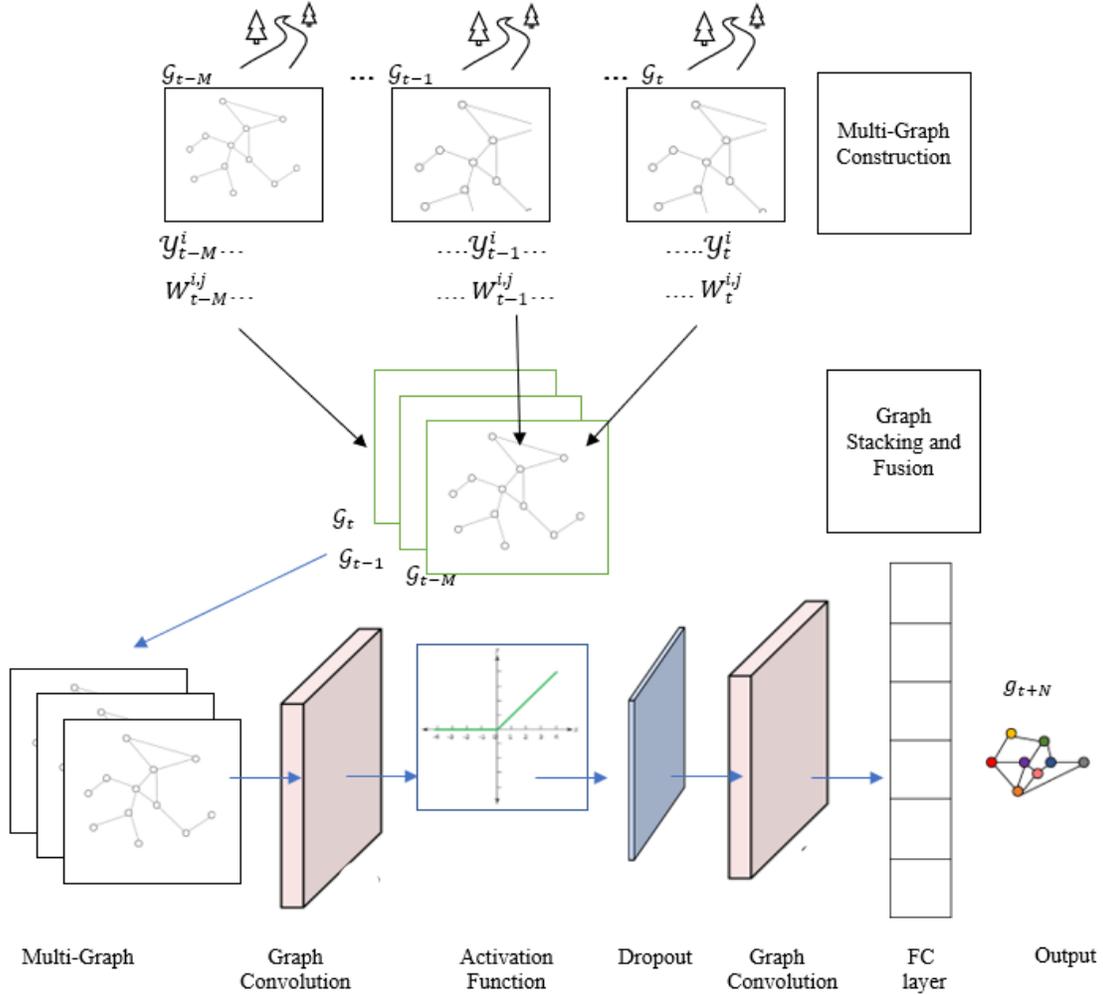

**Figure 1: Proposed MGCNN architecture for traffic flow prediction**

**Convolution on Graph**
A GCN normally operates on graph-structured data. Through consecutive message passing, GCN aggregates information from node to node and updates node representation based on edge information and the updated node information. Through these updates, every node on the graph is made aware of the information on adjacent nodes until every node is represented by information that takes into consideration the state of all other nodes on the graph. This process extracts the spatial correlations between the various nodes on the graph. The GCN convolutes through the constructed multigraphs to extract such spatial correlations over a desired time window. In GCN, two major approaches could be applied to perform the convolution: the spatial approach and the spectral approach. In this study, we adopt the spectral convolution method which allows for application of convolution on a spectral domain in large scale graphs [28]. In this approach, the Laplacian Matrix $L$ is used to represent the graph. Analyzing $L$ and its Eigen values helps to understand the graph structure [2]. The graph Laplacian matrix can be represented as follows:

$$L = D - W, \tag{2}$$

The normalized form of the graph Laplacian matrix is represented as:

$$L = I_n - D^{-\frac{1}{2}} W D^{-\frac{1}{2}}, \tag{3}$$

where $I_n$ is the identity matrix, $D \in \mathbb{R}^{n \times n}$ is diagonal degree matrix composing node degrees, $W$ is the adjacency matrix. Since our architecture constructs a graph for every timestep, the graph Laplacian matrix can be rewritten as:

$$L_t = I_n - D^{-\frac{1}{2}} W_t D^{-\frac{1}{2}}, \tag{4}$$

After analyzing the graph Laplacian $L_t$, a graph Fourier transforms a signal $z_t$, for every graph constructed at time $t$, which is defined as $z_t: \mathcal{V} \to \mathbb{R}$ to $\hat{z}_t = U^T z_t$, where $U$ is the graph Fourier basis This transformation enables the use of filtering operations in the convolution layer [32]. Accordingly, the spectral filtering operation is defined by a kernel $\phi$ and a convolution operator $*g$ as follows

$$\phi * g z_t = \phi(L) z_t = \phi(U \wedge U^T) z_t = U \phi(\wedge) U^T z_t \tag{5}$$

For large-scale graph analysis, it will be expensive enough to use this Laplacian decomposition directly. Thus, we apply a Chebyshev polynomial approximation to reduce parameters as the truncated expansion of $K - 1$ order where $K$ is the kernel size of the graph convolution. The graph convolution then can be expressed as

$$\phi * g z_t = \phi(L_t) z_t \approx \sum_{k=0}^{K-1} \theta_k T_k(\tilde{L}_t) z_t, \tag{6}$$

where, $T_k(\tilde{L}_t)$ is the Chebyshev polynomial of order $k$, $\tilde{L}_t = 2L_t/\lambda_{max} - I_n$, where $\lambda_{max}$ represents the largest eigenvalue of $L_t$. The convolution operator thus can be used in a multidimensional tensor form for a signal with desired input and output channel with the desired size of restricted localized filter [1]. The GCN layer can be applied to convolute the bidirectional multigraphs and predict flow data. We use this desired filtering size as per our lookback window $M$ to slide through the features of the multigraphs to convolute and extract meaningful spatial correlation. Thus, the model learns to predict traffic flow from a spatiotemporal perspective.

**Fully Connected Layer**
The output from the GCN is passed into a fully connected layer to predict the desired traffic states at the various nodes on the graph. In essence, the traffic prediction problem herein can be described as: given the graphs $\mathcal{G}_t = (\mathcal{V}, \mathcal{E}, W_t)$ at $m$ number of timesteps, our model $\mathcal{F}$ combines a sequence of graphs $[\mathcal{G}_{t-M}, \ldots \mathcal{G}_{t-3}, \mathcal{G}_{t-2}, \mathcal{G}_{t-1}, \mathcal{G}_t]$ of size $m$, to predict a graph structure $g_{t+N}: g_{t+N} \subseteq \mathcal{G}_{t+N}$ at timestep $t + N$ into the future, where $N$ is the prediction horizon. This can be represented as:

$$g_{t+N} = \mathcal{F}([\mathcal{G}_{t-M}, \ldots \mathcal{G}_{t-3}, \mathcal{G}_{t-2}, \mathcal{G}_{t-1}, \mathcal{G}_t]) \tag{7}$$

where, $M$ represents the lookback window used to capture historical temporal patterns in the graph states, and $m$ is the number of graphs over that lookback window which is defined based on the time step size. In this study, $g_{t+N}$ is defined based on the traffic state parameter of interest, which is a vector expressing turning movements at intersections after a predefined time $N$ into the future (also referred to as the prediction horizon). In $g_{t+N}$, the information at any node $v$ is defined as $x_{t+N}$, where $x = (x_1, x_2, x_3, \ldots x_s)$, where $s$ is the number of turning movements over that node. The size of the array $x_{t+N}$ defines the number of nodes in the fully connected layer.

**EXPERIMENTAL ANALYSIS**

**Study Area**
The proposed model is applied on Martin Luther King Blvd, an arterial in downtown Chattanooga, TN. As shown in Figure 2, we focus on a 1.2-mile segment extended from Central Avenue to Chestnut Street, which covers the busiest segment of MLK. The corridor has several signalized and unsignalized intersections. In this study, we focus on short-term prediction of turning movements at 10 signalized intersections along the corridor. For the graph representation, each intersection along the corridor is modeled as a node in the MGCNN, while the roadway links between intersections are modeled as edges. This graph representation of the corridor is shown in Figure 3.

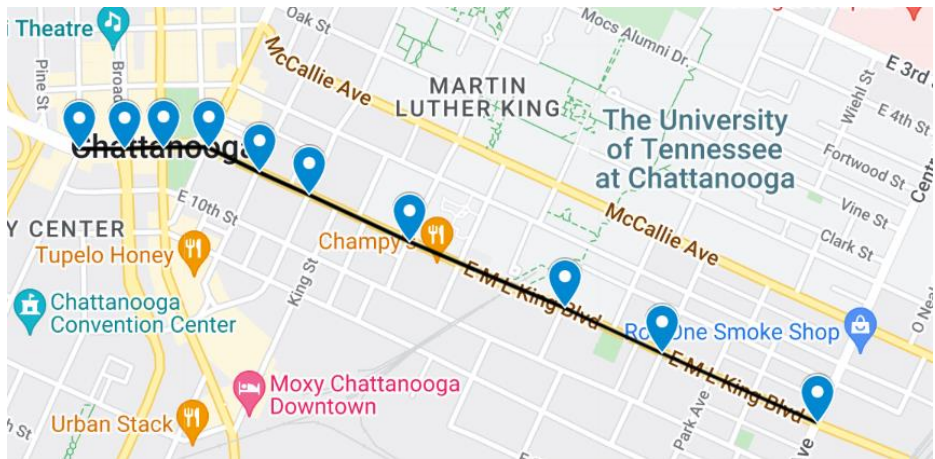
**Figure 2: EMLK Blvd Corridor**

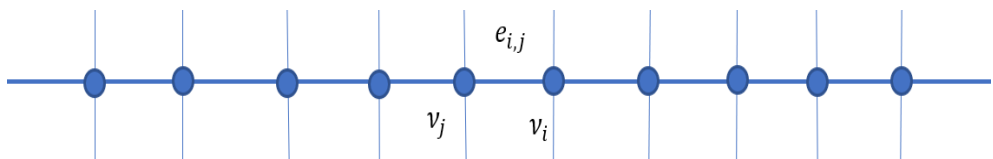
**Figure 3: Graphical representation of MLK corridor**

In our MGCNN, the intersections along the corridor are modeled as the in the graph structure and are vertices expressed by $\mathcal{V} = \{\mathcal{V}_1, \mathcal{V}_2, \ldots \mathcal{V}_n\}$. The edges in the model, $e_{i,j}$, represent the roadway link between intersections $i$ and $j$. The set $\mathcal{E}$ represents the edge structures between all intersections along the corridor. To form the weight matrix at any given time step t, travel times along the links were calculated using the speed data collected from GRIDSMART detectors along the corridor from each intersection. This is expressed as $w_t^{i,j} = travel\ time = distance\ betwen\ i\ \&\ j / speed\ at\ e_{i,j}$ for $i \rightarrow j$ direction, and $w_t^{j,i} = travel\ time = distance\ betwen\ i\ \&\ j / speed\ at\ e_{j,i}$ for the opposite direction.

**Data**

Data used in this study is collected from GRIDSMART cameras that are installed at every signalized intersection along the corridor. In accordance with their respective signal phases, data is collected in real-time for a variety of traffic performance parameters, including speeds, turning movement volumes, arrival on green, and many more. We used data from the month of June 2021's final 20 days for this study. A snapshot of the real-time data is shown in Table 2 and initially considered performance variables that are considered in this study are presented in Table 3.

**Table 2 Snapshot of Real-Time Data**

| Minute | Left Turn Count | Right Turn Count | Arrival on green for left turns | Arrival on red for left turns | Left Turn Speed (mph) | Thru Speed (mph) | Right Turn Speed (mph) | Green time for left turns (s) | .. |
|---|---|---|---|---|---|---|---|---|---|
| 0 | 4 | 0 | 1 | 2 | 25 | 35 | 15 | 4 | . |
| 1 | 2 | 0 | 0 | 2 | 30 | 40 | 35 | 3 | . |
| .. | 3 | 0 | 2 | 0 | 35 | 35 | 25 | 4 | . |
| 28800 | 4 | 0 | 3 | 1 | 35 | 35 | 15 | 6 | . |

**Table 3 Variable Description**

| Parameter | Unit | Description |
|---|---|---|
| **Traffic Count** | Numeric | Number of vehicles coming to the intersection. |
| **Speed** | MPH | Speed of the vehicle coming to the intersection. |
| **Arrival at green** | Numeric | Number of vehicles that are arriving at intersection when the signal is green. |
| **Arrival at red** | Numeric | Number of vehicles that are arriving at intersection when the signal is red. |
| **Arrival at yellow** | Numeric | Number of vehicles that are arriving at intersection when the signal is yellow. |
| **Green Time** | Second | Duration of the signal condition when it is green. |
| **Red Time** | Second | Duration of the signal condition when it is red. |
| **Occupancy time** | Second | Duration for which vehicles are occupying the road. |
| **Green occupancy** | Second | Duration for which vehicles are occupying the road while the signal is green |
| **Red occupancy** | Second | Duration for which vehicles are occupying the road while the signal is red. |
| **Yellow occupancy** | Second | Duration for which vehicles are occupying the road while the signal is yellow. |
| **Class** | 0 or 1 | A categorical variable where 0 indicates weekend and 1 indicate weekday. |

The real time data in Table 2 is built upon GRIDSMART data and arranged lane by lane from North, South, and West bounds. All four bounds, each having 3 possible turning movements

*Palit and Osman*

(right, left, and thru), result in 12 different lane-bound combinations a shown in Figure 4. Thus, for first 11 measures described in Table 3, 132 possible features are formed which are expressed in Table 2 and it is repeated for all ten intersections along the corridor and for 20 consecutive days. Each of the performance measures (features) in Table 3 are collected for every leg in every intersection forming a vector set of $133 \times 1$ which are denoted in the proposed model as $\mathcal{Y} = (\mathcal{Y}_1, \mathcal{Y}_2, \mathcal{Y}_3, \ldots \mathcal{Y}_\mathcal{K})$ in the graph at each time step $t$. Since the data is minute-by-minute, the data at every intersection can be represented by a $133 \times 28800$ tensor, where consecutive 20-day minute by minute forms 28800 in length and 133 represent all features in the data. For each vertex $\mathcal{V}_i$, this feature tensor is denoted as $\mathcal{Y}_t^i$ in the model. We aim to predict 12 directional turning movements, $(x_1, x_2, x_3, \ldots x_s)$ at every intersection in $N$ prediction horizon into the future. This output array is expressed by $x_{t+N}$ which provides the information on the subset graph $g_{t+N}$.

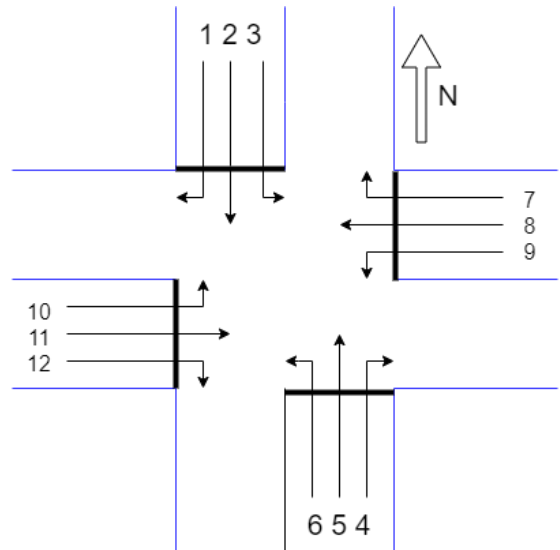

**Figure 4: 12 Possible Turning Movements at an Intersection**

**Data Insights**

To extract statistical insights, locations statistics of 12 possible turning movement of Chestnut intersection is shown in Figure 5.

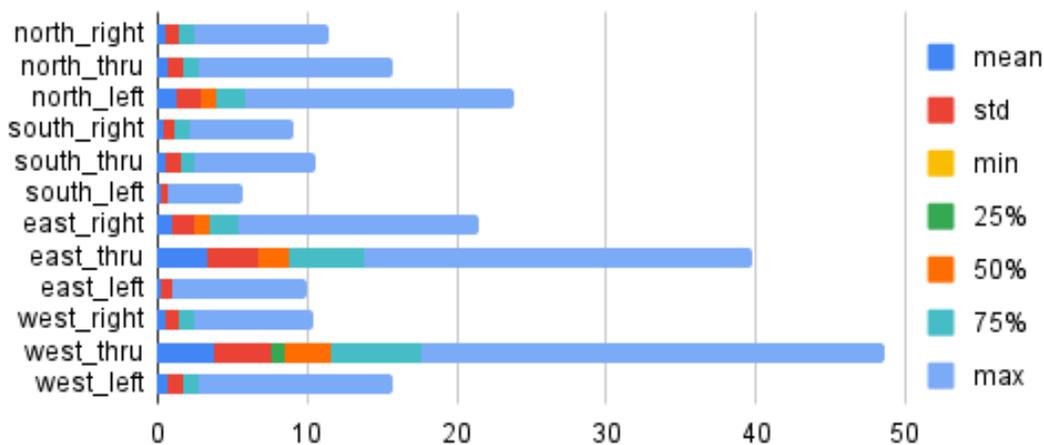

**Figure 5: Location Statistics of Turning Movement Counts**

The locations statistics provide some useful insights. We can specially notice the absence of 25 % and minimum data points on the diagrams referring to the fact that 0 vehicle is both the minimum and 25th percentile. The standard deviation is greater in traffic counts for Eastbound (EB) and Westbound (WB) thru movements than others, reflecting more dispersion in these movements and more spread-out data. These two movements also contain the highest peak volume and highest means as seen from the mean and maximum. Westbound thru movement carries more traffic than other main MLK corridor movements (EB-WB) which is evident from one day volume at Chestnut (12th June) represented in Figure 6. We also clearly notice two peaks: morning peak around 10 AM to 2 PM and evening peak around 5 PM to 9 PM.

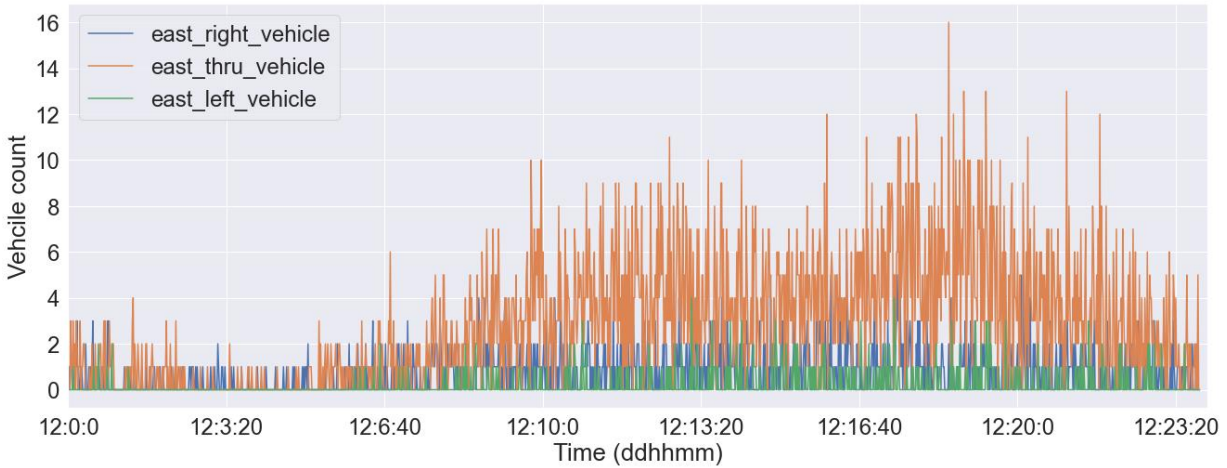

(a)

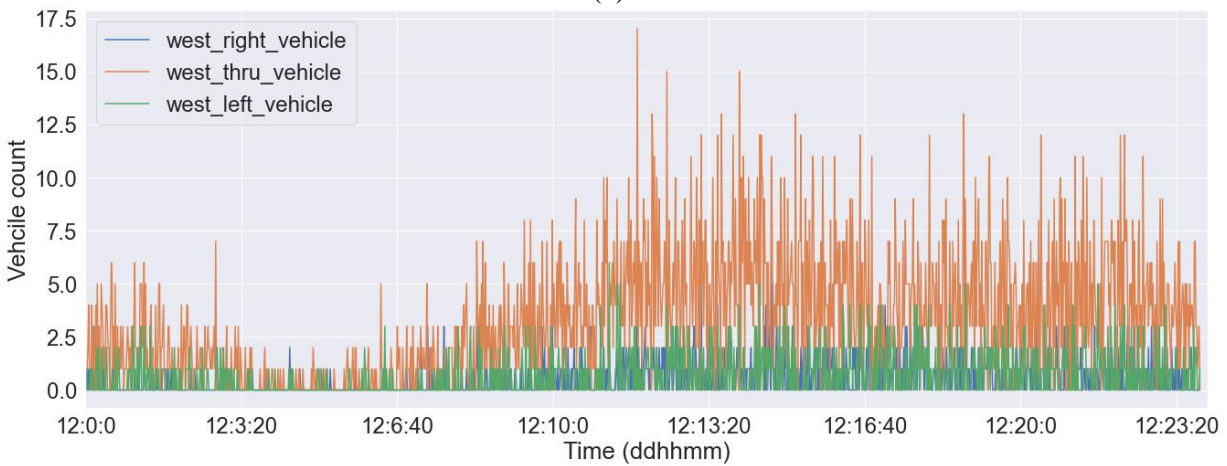

(b)

**Figure 6: Eastbound & Westbound Traffic Count at Chestnut Street**

**Dimensionality Reduction**

Each vertex $\mathcal{V}$ of the graph, represented by each intersection consists of 133 attribute which is comparatively large in volume and can cause exhausting computation. Different data analytic techniques are applied to reduce the attribute size. Minute by Minute data can contain large number of smaller values. Attributes that contain lots of 0 values are mainly occupancy time: green, red, and yellow occupancy time. While MLK is a busy corridor and largest share of signal state is in green, red and yellow occupancy time contribute less due to large amount of 0 values. During green, vehicle do not tend to stop hence resulting in zero occupancy time. Thus, occupancy time, green occupancy, red occupancy, and yellow occupancy time do not contain significant information. We drop these 4*12= 48 attributes to reduce volume of the data which leads us to 85 attributes which we annotate from A1 to A85 as shown in the Table 4. For the existing attributes we check multicollinearity to further filtering our attributes. The correlation heatmap in Figure 7 produced for Chestnut intersection provide us with decisions to further filtering out less important attributes. The heatmap provide us some important insights into the relationship between attributes. We can see almost all predictors are well related to the target variable i.e., traffic count (A1 to A12). The predictors have some highly correlated clusters among them which we focus for. For example, we can see attributes of A15 to A19 are correlated with attributes of A63 to A67. We notice similar correlations between A13 to A22 and A49 to A58; A52 to A59 and A64 to A71. One of the strongly correlated clusters remain between attributes of A25 to A36 and A37 to A48. Collinearity of more than 0.8 to 0.9 is alarming [34]. Thus, we keep only one of the highly correlated (0.8 to 0.9) attributes between a pair and dispose the other. For instance, we dispose attributes A25 to A36, A13, A14, A16, A17, A18, A54, A55, A56, A58 and A59. This result in 63 final attributes to be considered as input.

**Table 4: Attribute Annotation**

| Parameter | Annotation |
|---|---|
| **Traffic Count** | A1 to A12 |
| **Speed** | A13 to A24 |
| **Arrival at green** | A25 to A36 |
| **Arrival at red** | A37 to A48 |
| **Arrival at yellow** | A49 to A60 |
| **Green Time** | A61 to A72 |
| **Red Time** | A73 to A84 |
| **Class** | A85 |

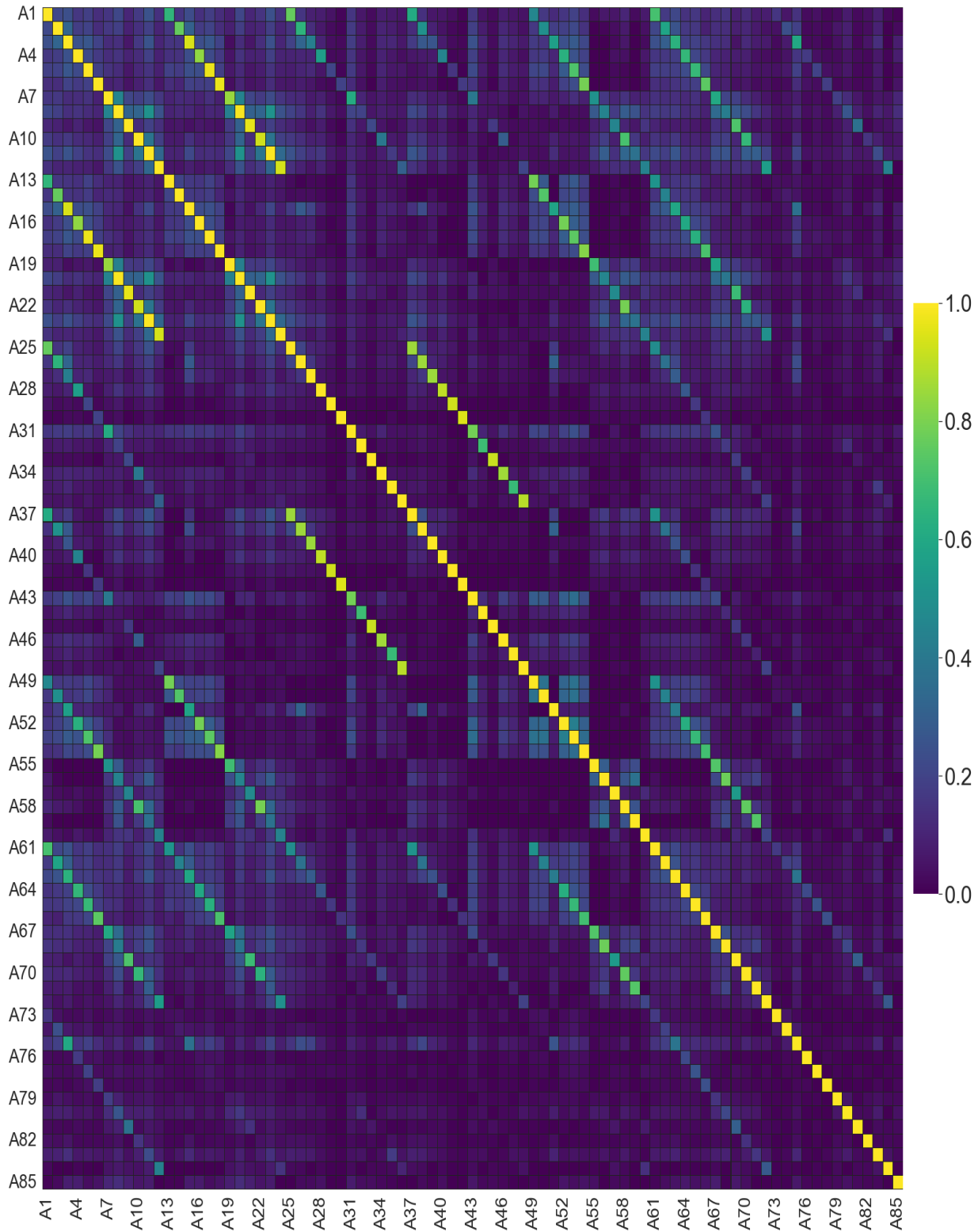

**Figure 7: Correlation Heatmap**

**Handling Outlier**

Outlier detection is critical for good data mining and machine learning applications since it improves data quality. Outliers are defined as extremely big or small values compared to the data range, which might cause the data to deliver erroneous information and impact prediction outcomes in machine learning challenges. One popular way to detect outlier is based on boxplot and interquartile range (IQR) [35]. For a given range of data, if first quartile (25%) is denoted as Q1, third quartile (75%) is denoted as Q3, the interquartile range, $IQR = Q3 - Q1$. This applies with lower bound= $Q1 - 1.5 \times IQR$ and upper bound= $Q3 + 1.5 \times IQR$. Any data points outsides these bounds are detected as outlier as shown in Figure 8(a) and needs to be treated. Through boxplot of our target attributes A1-A12 (turning volume) for Chestnut intersection in Figure 8(b) we can clearly see the outliers outside the bounds which we replace with median[36] of the data resulting in finely prepared dataset ready to be used as MGCNN model input. We adopt similar strategy for all intersections accordingly. Thus, after dimensionality reduction and outlier analysis, our model input dimension changes from $133 \times 28800$ tensor to $63 \times 28800$ tensor as 133 attributes are reduced and refined to 63 that play critical roles.

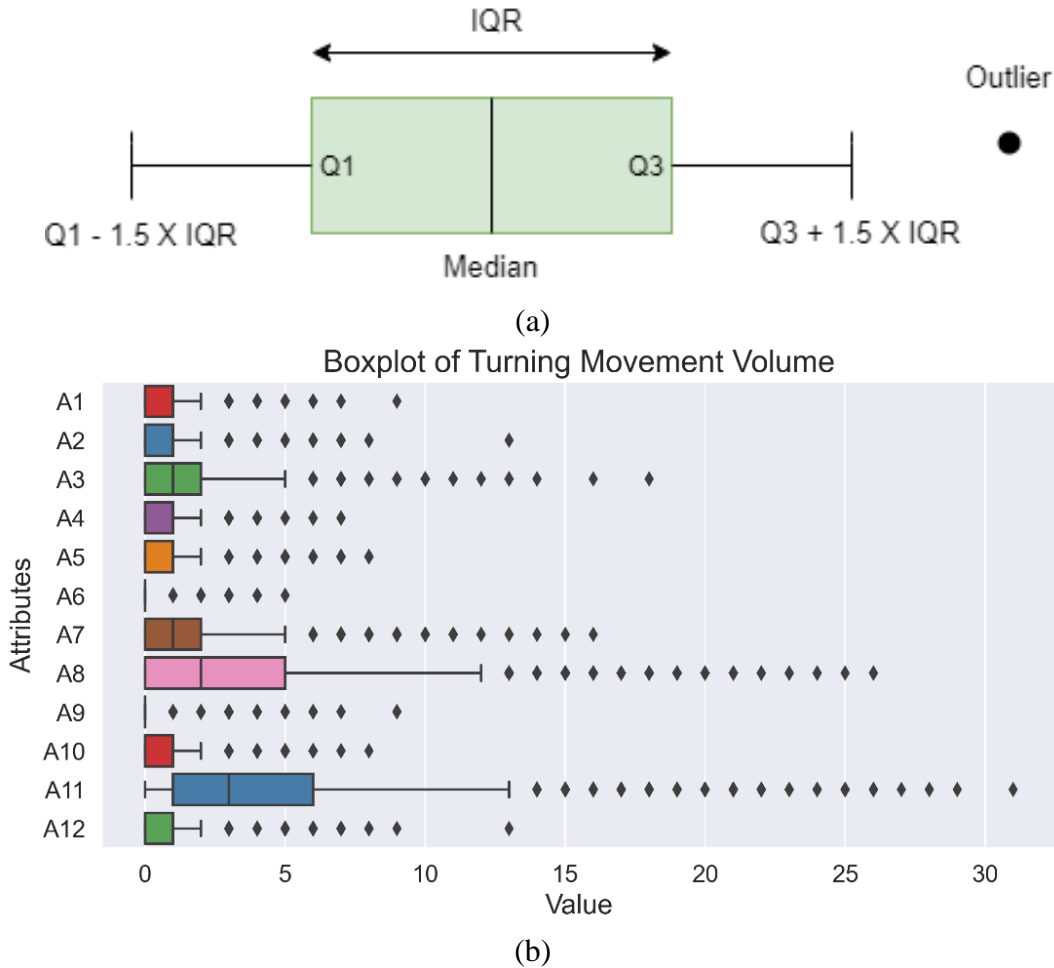

(a)

(b)

**Figure 8: Outlier Detection and Handling Strategy**

*Palit and Osman***Evaluation Metrics and Baseline**

To evaluate our model and measure its performance, we adopt 4 different performance measures: Mean Squared Error (MSE), Root Mean Square Error (RMSE), Mean Absolute Error (MAE), and mean absolute percentage error (MAPE). Each evaluation metric comes up with its advantages which can help provide a better assessment of the model performance. These measures are defined as follows:

(i) Mean Square Error (MSE) is one of the widely used metrics which combines the measurement of bias and variance of prediction values. A model with lower MSE is one with better performance. MSE is calculated by

$$\text{MSE} = \frac{1}{N}\sum_{i=1}^{N}(y_{true} - y_{predicted})^2 \tag{8}$$

(ii) Root Mean Square Error (RMSE) is the root of MSE. It usually reflects the performance of a model better and measures the spread of errors. Lower values mean better performance. It is expressed as

$$\text{RMSE} = \sqrt{\frac{1}{N}\sum_{i=1}^{N}(y_{true} - y_{predicted})^2} \tag{9}$$

(iii) Mean Absolute Error (MAE) is a simple and widely used measure. However, less biased toward higher values and not sensitive enough to the outliers. It is expressed as

$$\text{MAE} = \frac{1}{N}\sum_{i=1}^{N}|y_{true} - y_{predicted}| \tag{10}$$

(iv) Mean Average Percentage Error (MAPE) overcomes the limitations of MAE and normalizes the errors. Unlike other measures, it is expressed in percentages. MAPE can be calculated by

$$\text{MAPE} = \frac{100\%}{N}\sum_{i=1}^{N}\left|\frac{y_{true} - y_{predicted}}{y_{true}}\right| \tag{11}$$

We compare our model to four baseline state-of-the-art models to evaluate where MGCNN stands. These methods are: 1) Temporal graph convolutional network (T-GCN) [22], 2) Diffusion Convolutional Neural Network (DCRNN) [20], 3) Traditional Gated Recurrent Unit (GRU) [18] and 4) Traditional Graph Convolution with Multilayer Recurrent Neural Network (GCN+RNN) [2]. First, we perform a sensitivity analysis to identify the optimal lookback window, then the short-term prediction performance is evaluated for five *N* values (1, 2, 3, 4, and 5 min).

**Experimental Result**

Our experiments were conducted on Two 14-Core Intel Xeon E5-2680 V4 Processors, with a 128 GB Ram, and an Nvidia P100 GPU (16GB) with 1792 double-precision cores. As discussed earlier, our model MGCNN operates with two GCN layers, one drop-out layer, connected with a fully connected output layer. A summary of the model hyperparameters is provided in Table 5. The model was trained with MSE as the loss function and using first 19 days of the 20-day data, while the remaining one-day data was used for testing.

**Table 5: Model Summary**

| Hyperparameter | Value |
| --- | --- |
| **Learning rate (Lr)** | 0.0007 |
| **Learning Rate Decay** | 0.1 by 10 step |
| **Optimizer** | Adam |
| **Dropout rate** | 35% |
| **Batch Size** | 16 |

We first train our model considering six different lookback window sizes varying from 10 minutes to 60 minutes with a 10-minute increment and a 5-minute prediction horizon. The training performance depicted in Figure 9 shows that the model performance stabilizes fast at a loss value of around 1.0 after about 10 epochs regardless of the lookback window size while lookback window of 10 min shows the lowest error. Further investigation into the model performance at the different prediction horizons shows that a 10-minute lookback can provide slightly better performance compared to other values as shown in Table 6. Thus, we continue the analysis of the model performance considering the 10-minute lookback window as the optimum window.

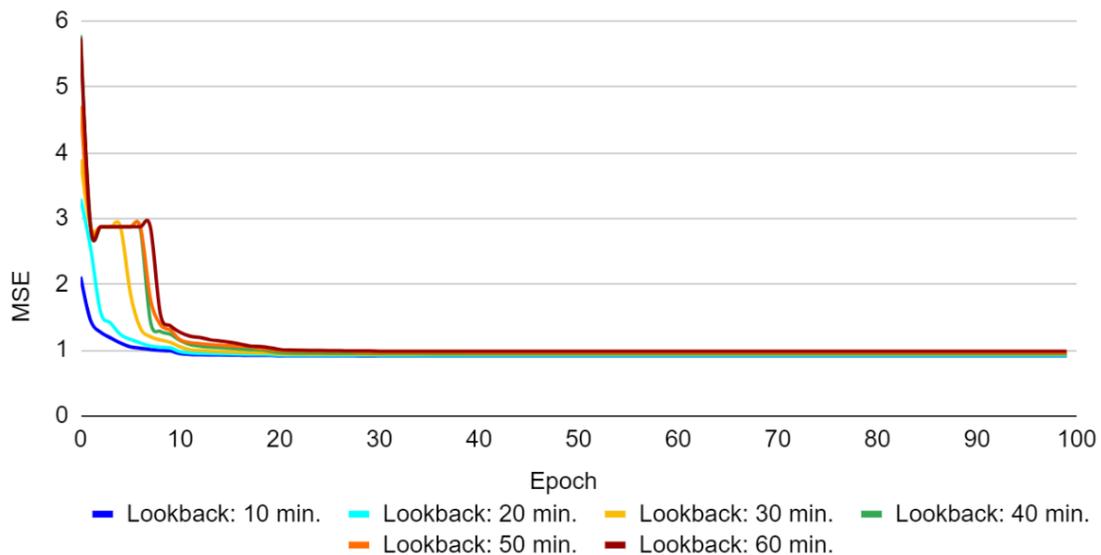

**Figure 9: Training Curves for different lookback windows over 5-minute horizon**

**Table 6: Test results over prediction horizon: 5-minute**

| Lookback Window(min) | MSE | RMSE | MAE | MAPE |
| --- | --- | --- | --- | --- |
| 10 | 0.985 | 0.908 | 0.512 | 4.205% |
| 20 | 0.995 | 0.913 | 0.514 | 4.238% |
| 30 | 1.005 | 0.917 | 0.518 | 4.251% |
| 40 | 1.014 | 0.921 | 0.521 | 4.268% |
| 50 | 1.037 | 0.932 | 0.529 | 4.315% |
| 60 | 1.052 | 0.937 | 0.532 | 4.325% |

To enable applications such as adaptive signal control and corridor management, short-term traffic prediction over shorter time periods is needed. Therefore, we test the model performance to predict turning movements over five prediction horizons ranging from 1 minute to 5 minutes with a 1-minute increment. The training performance of the model depicted in Figure 10(a) shows that the model converges after around 10 epochs with a training MSE of around 0.9 for all prediction horizon values. This performance indicates the model's ability to learn the spatiotemporal correlations fast. The multigraph structure along with the GCN that convolutes through these graphs over time facilitates quick capturing of how each intersection affects the other intersections along the corridor, and how the traffic patterns change over short time periods.

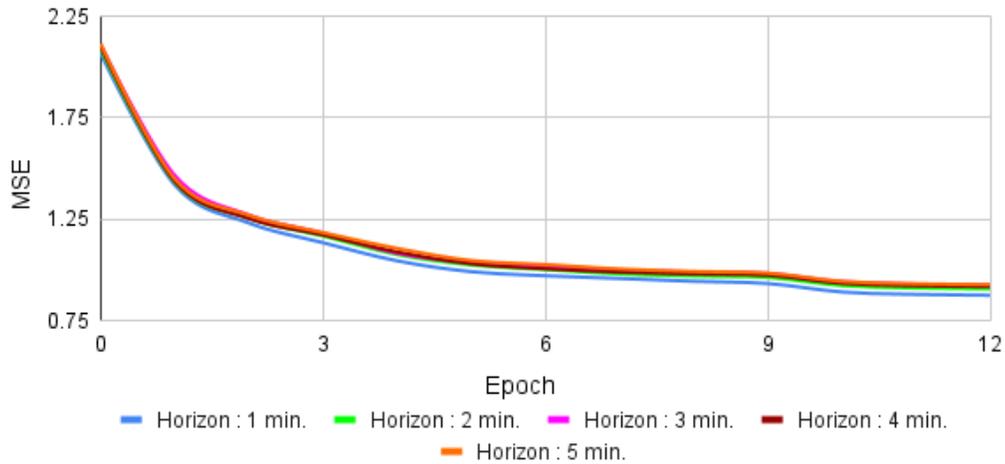

(a)

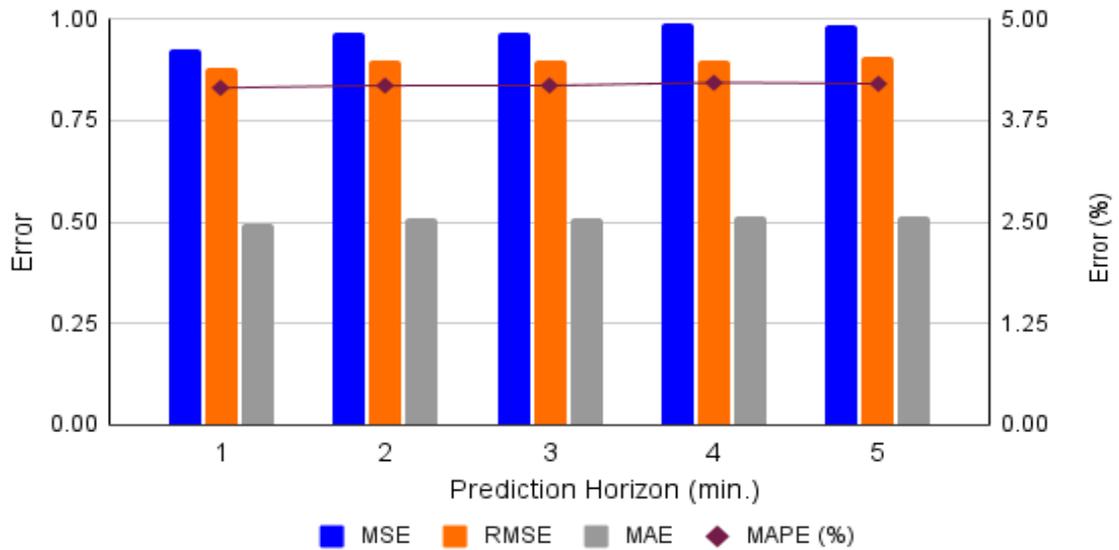

(b)

**Figure 10: Model performance in lookback window :10 minute -(a) Training Loss (b) Test Errors**

Similarly, the testing performance of the model shows a balanced and stable prediction performance for all prediction horizons, as shown in Figure 10(b). For instance, the MSE value changes from 0.9252 for the 1-minute prediction horizon to 0.9874 for the 5-minute prediction horizon. Since targeting longer prediction horizons introduces more uncertainty and non-linearity in the data, the prediction results show the models ability to capture such temporal dynamics efficiently, hence presents the model as a viable enabler for proactive traffic management.

The efficiency of the model is further expressed through ground truth vs prediction plot of traffic count in Figure 11 from one major turning movement: East Thru as example. From the figure we can clearly see that the predicted and true values follow almost the same pattern thus enabling the model to capture the data efficiently as found in test results. This could come handy in traffic management application to optimize signal control system and control congestion.

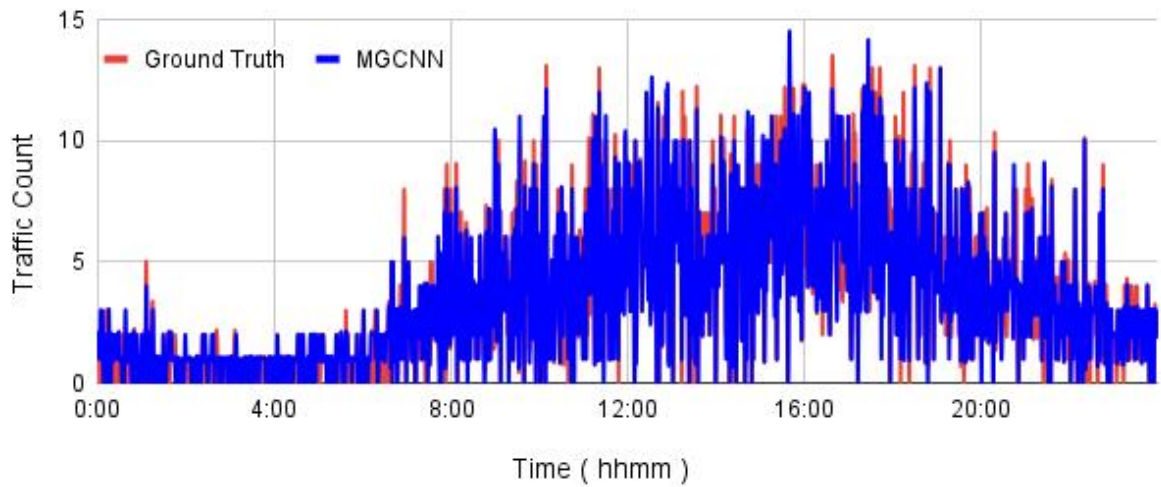

**Figure 11: Ground Truth vs Prediction plot for westbound thru movement: Chestnut**

To further evaluate the proposed model, we compare its prediction performance to four baseline state-of-the-art models: T-GCN, DCRNN, GRU, and GCN+RNN in Table 7. The comparative results in terms of the four selected performance metrics are presented in Table. The table clearly shows that all models provide relatively good results. However, comparative evaluation proves the superiority of the proposed MGCNN model with more accurate turning movement predictions over the various prediction horizon values. This is supported by the MSE, RMSE, MAE and MAPE values in the table. For other models, when MSE is about 2, MGCNN drops to roughly 0.9. MGCNN has a better margin of loss than the other four state-of-the-art models on additional measures too such as RMSE, MAE and MAPE. While the results show that all models have stable performance moving from one prediction horizon to the other, the MGCNN model's performance is yet comparatively better as the resulting absolute errors are less. Overall, the results prove the superiority of the proposed MGCNN model compared to the other baseline models in capturing spatiotemporal correlations along the test transportation corridor. We can also see that as we progress from smaller prediction horizons to bigger ones, the errors increase. It makes sense because as the time horizon increases, traffic dynamics vary more, adding more unpredictability and non-linearity.

**Table 7: Performance comparison with respect to established & traditional approaches**

| Model | Performance Metrics | *Prediction Horizon (N)* | | | | |
| --- | --- | --- | --- | --- | --- | --- |
| | | 1-min | 2-min | 3-min | 4-min | 5-min |
| DCRNN | MSE | 2.2257 | 2.2266 | 2.2268 | 2.2274 | 2.2278 |
| | RMSE | 1.3290 | 1.3290 | 1.330 | 1.3340 | 1.3370 |
| | MAE | 0.6860 | 0.680 | 0.6840 | 0.6870 | 0.6890 |
| | MAPE | 4.9550% | 4.9680% | 4.9790% | 4.9710% | 4.9830% |
| T-GCN | MSE | 2.1055 | 2.1422 | 2.1731 | 2.2018 | 2.2021 |
| | RMSE | 1.2624 | 1.262 | 1.2621 | 1.2625 | 1.2627 |
| | MAE | 0.6584 | 0.6605 | 0.6700 | 0.6729 | 0.6746 |
| | MAPE | 4.3560% | 4.3590% | 4.3603% | 4.3660% | 4.3710% |
| GRU | MSE | 2.0592 | 2.0598 | 2.0603 | 2.0609 | 2.0609 |
| | RMSE | 1.2455 | 1.2468 | 1.2475 | 1.2479 | 1.2478 |
| | MAE | 0.6652 | 0.6645 | 0.6661 | 0.6664 | 0.6669 |
| | MAPE | 4.275% | 4.2752% | 4.2757% | 4.2758% | 4.2760% |
| GCN+RNN | MSE | 2.0531 | 2.0574 | 2.0593 | 2.0607 | 2.0613 |
| | RMSE | 1.2469 | 1.2499 | 1.2504 | 1.2522 | 1.2528 |
| | MAE | 0.6471 | 0.6483 | 0.6505 | 0.6529 | 0.6546 |
| | MAPE | 4.1879% | 4.1960% | 4.2030% | 4.2110% | 4.2170% |
| **MGCNN** | **MSE** | **0.9252** | **0.9661** | **0.9670** | **0.9907** | **0.9874** |
| | **RMSE** | **0.8810** | **0.9040** | **0.8990** | **0.9010** | **0.9081** |
| | **MAE** | **0.4970** | **0.5070** | **0.5070** | **0.5130** | **0.5128** |
| | **MAPE** | **4.1554%** | **4.1831%** | **4.1840%** | **4.2186%** | **4.2140%** |

Figure 12 demonstrates the comparison between performances of state-of-art models with MGCNN based on one-hour westbound thru traffic count. We can see all models capture the trend well. However, if we look closely, we can see MGCNN still captures the ground truth volumes more accurately than others. GRU and hybrid model traditional GCN+GRU in Figure 12 (c) & (d) show close performance accuracy as evident from Table 7.

All the baseline models rely on hybrid architectures with two major components: one aiming to capture temporal correlations (e.g., GRU and RNN) and another aiming to capture spatial correlations (e.g., CNN). Having such hybrid architectures over a graph adds to the complexity of these models. On the other hand, our proposed MGCNN captures the temporal correlations through the constructed multiple graphs, while the spatial correlations are captured through the CNN component of the model. This relatively simpler architecture (compared to the other models) is evident in the training speed of the different models as the TGCN, GCN+RNN, and GRU models require around 60-80 minutes to train on the six-data. The DCRNN model requires more time of around 60-90 minutes to train, while our MGCNN model requires around 25-30 minutes to train on the same data. This indicates how complex are the architectures of the other models which makes them exhaust to train compared to the MGCNN model. With our model achieving higher prediction accuracies, the multigraph fusion with the spectral convolution proves superior to state of art models for short-term turning movement predictions.

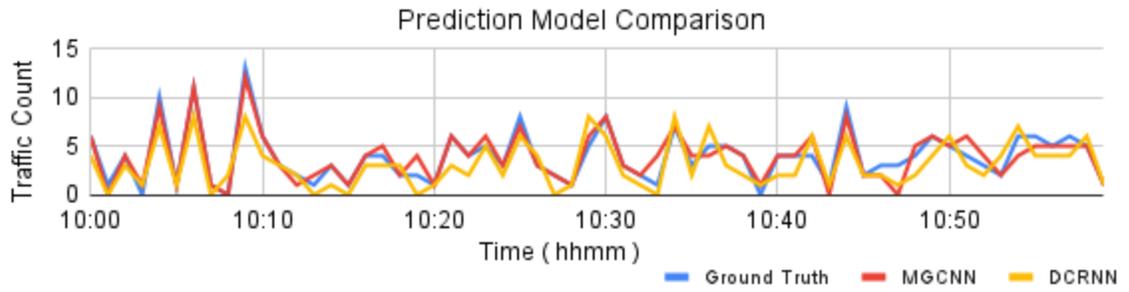

(a)

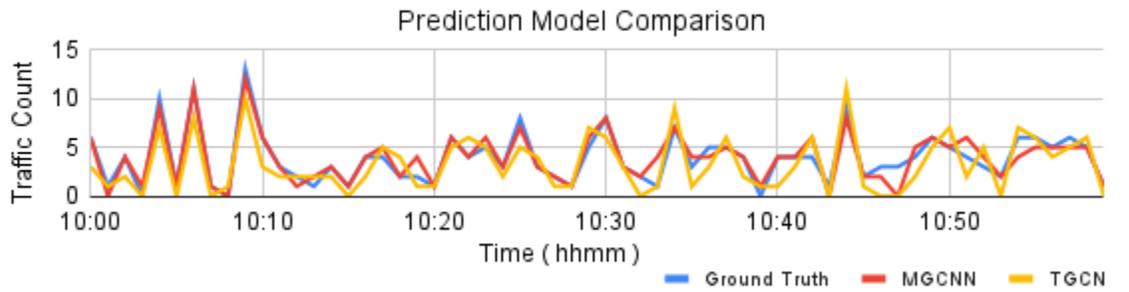

(b)

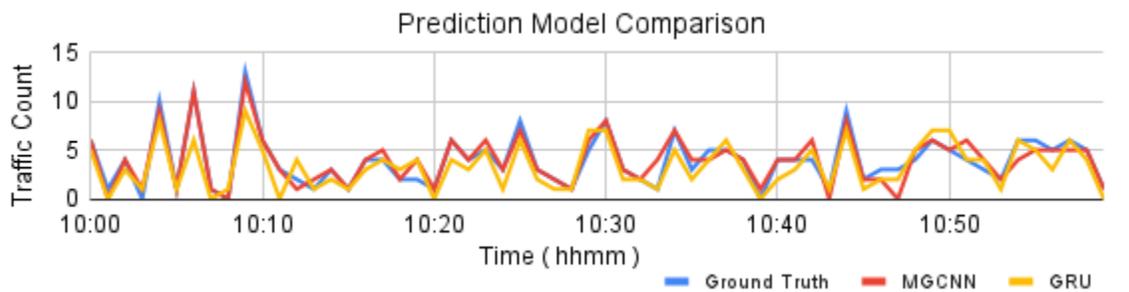

(c)

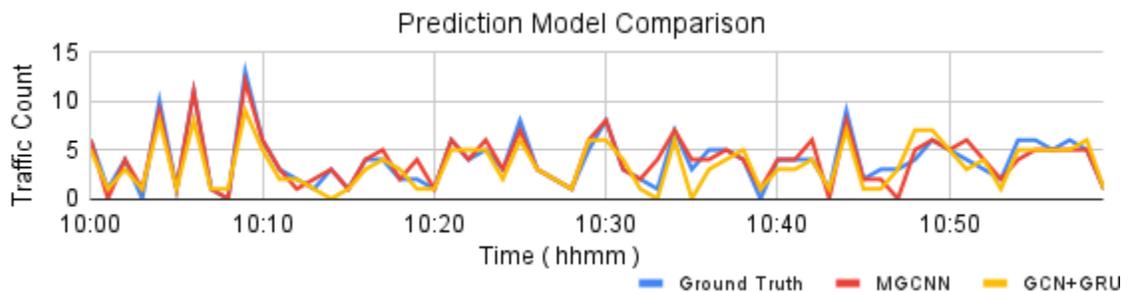

(d)

**Figure 12: Performance Comparison of State-Of-Art Models with MGCNN**

.

*Palit and Osman*

**CONCLUSION**

This study introduces a deep learning architecture, referred to as a multigraph convolution neural network (MGCNN) that combines a multigraph structure with a spectral convolution to perform short-term prediction of turning movements at intersections. The proposed architecture captures the temporal correlations in traffic data through the constructed graphs, while understands the spatial correlations through the convolution over each graph. The model was tested for different values of the lookback window to identify the optimal value for further analysis. The model prediction performance was then tested for five prediction horizon values ranging from 1 to 5 minutes. The reason for targeting short prediction horizons is to test the model's ability to capture minimal changes over a short time frame. Finally, the model's performance was evaluated against four baseline state-of-the-art models to verify its effectiveness.

The sensitivity analysis for the lookback window shows that a 10-minute historical look at the data provides the best prediction results 5 minutes into the future compared to longer lookback windows. This indicates that further back historical data may not be relevant to conditions into the future. Additionally, the results showed that the model is capable of accurately predicting turning movements at the various intersections with MSE value ranging from 0.9252 to 0.9874 for prediction horizons 1 to 5 minutes, respectively. These results were significantly better than those of the baseline models which showed MSE values around 2.2257 for the DCRNN model, 2.1055 for the T-GCN model, 2.0592 for the GRU and 2.0531 for GCN+RNN model. While the latter baseline may seem to provide comparative results to our model in terms of MAPE, its complexity showed in the training which required close to twice the time to converge compared to our proposed model. More so, when investigating the training speeds of the other models, they were all at least 2 times that of our proposed model.

The presented results show that the proposed model predicts multi-directional traffic flow over short prediction horizons efficiently, proving the model's ability to capture minimal changes in traffic conditions. The proposed architecture enabled capturing spatial-temporal correlations in traffic data and modeling spatial-temporal traffic dynamics efficiently. While the other more complex models' performance could yet be considered acceptable, our proposed architecture is capable to significantly minimize the prediction errors with the minimal training cost possible. Effective application of this model in the real field can prove a wide range of functionality for proactive traffic management and signal control at intersections.

**ACKNOWLEDGMENTS**

This material is based upon work supported by the Department of Energy, Office of Energy Efficiency and Renewable Energy (EERE), under Award Number DE-EE0009208. Any opinions, findings, and conclusions, or recommendations expressed in this material are those of the author(s) and do not necessarily reflect the views of the Department of Energy.

      We would like to also acknowledge Mr. Austin Harris for his support in the data extraction effort from the Grid Smart cameras. We also would like to acknowledge Mr. Akintoye Oloko for the constructive discussions in the early stages of the work.

**AUTHOR CONTRIBUTIONS**

The authors confirm contribution to the paper as follows: study conception and design: O. A. Osman and J. R. Palit; data collection: O. A. Osman; analysis and interpretation of results: J. R. Palit and O. A. Osman; draft manuscript preparation: J. R. Palit and O. A. Osman. All authors reviewed the results and approved the final version of the manuscript.